\documentclass[conference]{IEEEtran}
\IEEEoverridecommandlockouts

\usepackage{cite}
\usepackage{amsmath,amssymb,amsfonts}
\usepackage{algorithmic}
\usepackage{graphicx}
\usepackage{textcomp}
\usepackage{xcolor}
\def\BibTeX{{\rm B\kern-.05em{\sc i\kern-.025em b}\kern-.08em
    T\kern-.1667em\lower.7ex\hbox{E}\kern-.125emX}}
\usepackage[firstpage=true]{background}
\begin{document}
\SetBgContents{\parbox{\textwidth}{\footnotesize © 2026 IEEE. Personal use of this material is permitted. Permission from IEEE must be obtained for all other uses, in any current or future media, including reprinting/republishing this material for advertising or promotional purposes, creating new collective works, for resale or redistribution to servers or lists, or reuse of any copyrighted component of this work in other works.}}
\SetBgScale{1}
\SetBgAngle{0}
\SetBgPosition{current page.north}
\SetBgVshift{-1cm}
\SetBgColor{black}
\SetBgOpacity{1}
\title{Point Cloud Upsampling through Patch-based Frequency Superposition\\
\thanks{This work was partly funded by the Deutsche Forschungsgemeinschaft (DFG, German Research Foundation) Project ID 532402151 and – SFB 1483 – Project-ID 442419336, EmpkinS. 
We also acknowledge funding received from the Bavarian Ministry of Economic Affairs, Regional Development and Energy within the scope of the funded project ”Bavarian Advanced Resolution Radar (BAVAR-RADAR)”, funding label DIK0622.}
}

\author{\IEEEauthorblockN{Marina Ritthaler, Azhar Hussian, Vasileios Belagiannis, André Kaup}
\IEEEauthorblockA{\textit{Chair of Multimedia Communications and Signal Processing} \\
\textit{Friedrich-Alexander-University Erlangen-Nuremberg}\\
Erlangen, Germany \\
marina.ritthaler@fau.de, azhar.hussian@fau.de, vasileios.belagiannis@fau.de, andre.kaup@fau.de}
}

\maketitle

\begin{abstract}
In recent years, neural networks have become the dominant models in most point cloud upsampling methods. Although these approaches are achieving good results, they do have drawbacks, such as a lack of interpretability and data dependency. Moreover, they have to be trained on a dataset that is similar to the test data in order to perform well. To avoid these disadvantages, we propose Point Cloud Upsampling through Patch-based Frequency Superposition (PUtPFS), an optimization-based approach that selects subsets of points and estimates the surface of this set through superpositioning spatial frequencies. Then, new points are placed on this surface. By successively selecting points in the least dense regions of the point cloud, a uniform upsampling can be reached.
With this method, we surpass the current best upsampling results in the commonly considered point-to-surface distance. Furthermore, we achieve the best Chamfer and Hausdorff distance among the optimization-based approaches. As an additional advantage, our method does not need any training data and is mathematically interpretable.
\end{abstract}

\section{Introduction}
\label{sec:intro}

\noindent With the increased availability of 3D sensors like LiDAR, three-dimensional data is often represented as point clouds. Point clouds are unordered collections of data points in space. Each point consists of coordinates in 3D space and can optionally have attributes such as color, intensity, or a label. Those point clouds are used in autonomous driving, industrial automation, or medical applications. When capturing the real-life environment, objects or object parts that are further away from the scanner are represented with fewer points and, therefore, at a low resolution. Point cloud upsampling is used to increase the resolution by adding more points to the point cloud. This is often a first step before other tasks are performed, such as 3D shape classification, 3D object detection, or 3D object segmentation. More uniform and dense point clouds lead to better results than using raw data \cite{reviewPaper}. For this densifying we propose our new method, which we term Point Cloud Upsampling through Patch-based Frequency Superposition (PUtPFS). It approximates local surface regions with a sparse superposition of parametric basis functions.

\section{Related Work}
\label{sec:relwork}

\noindent In recent years, many strategies have been developed to address the problem of densifying point clouds. Those methods can be divided into classical optimization-based approaches and deep learning-based methods.\\ 
\indent Since the release of PU-Net, deep learning-based methods have dominated the field. The most important building block concerns the extraction of features. While PU-Net \cite{PUNET} uses Pointnet++ \cite{Pointnet++} for feature extraction, others use graph convolution-based methods, e.g., PU-GAN \cite{PUGAN}, PU-GCN \cite{PUGCN}, GC-PCU \cite{GCPCU}. In Point Transformer \cite{PointTrans} local and global features are generated in two network branches and are related through an attention module. An alternative approach to the upsampling problem involves self-supervised methods. SPU-Net \cite{SPUNET} was developed with the objective of not needing ground truth dense point sets for the supervision. Recently, SPU-PMD \cite{SPUPMD} was published, which adds points through a self-supervised topological mesh deformation network.\\
\indent Although deep learning-based methods can achieve good results, they do exhibit some drawbacks. For one, the achieved results are only reached when the network is trained on the same type of data on which it is evaluated. When an upsampling method is required for a small dataset, where no training data of the same type is available, the upsampling results degrade. Additionally, neural network-based methods always have the risk that they hallucinate or distort the geometry if outliers are part of the input. Lastly, they are black-box approaches, which constitutes a problem for sensitive applications, for example, in the medical domain.\\
\indent Optimization-based upsampling techniques estimate the geometry using the L1-median, including LOP \cite{LOP}, WLOP \cite{WLOP}, CLOP \cite{CLOP}, and EAR \cite{EAR}. Among these, EAR is most widely used, as it further incorporates normals in an edge-aware manner to better preserve sharp features. An attempt to use frequency models to upsample the geometry of point clouds was made with FSGU \cite{FSGU}. Here, the point cloud is voxelized and every voxel is individually upsampled. Therefore, the variances of the three dimensions are calculated, and it is assumed that the dimension of the smallest variance portrays a non-closed function that represents a smooth surface. The authors then used a frequency model to estimate this function.\\

\section{PROPOSED ALGORITHM}
\label{sec:algo}
\begin{figure}[t]
\centering
\centerline{\includegraphics[height=14cm]{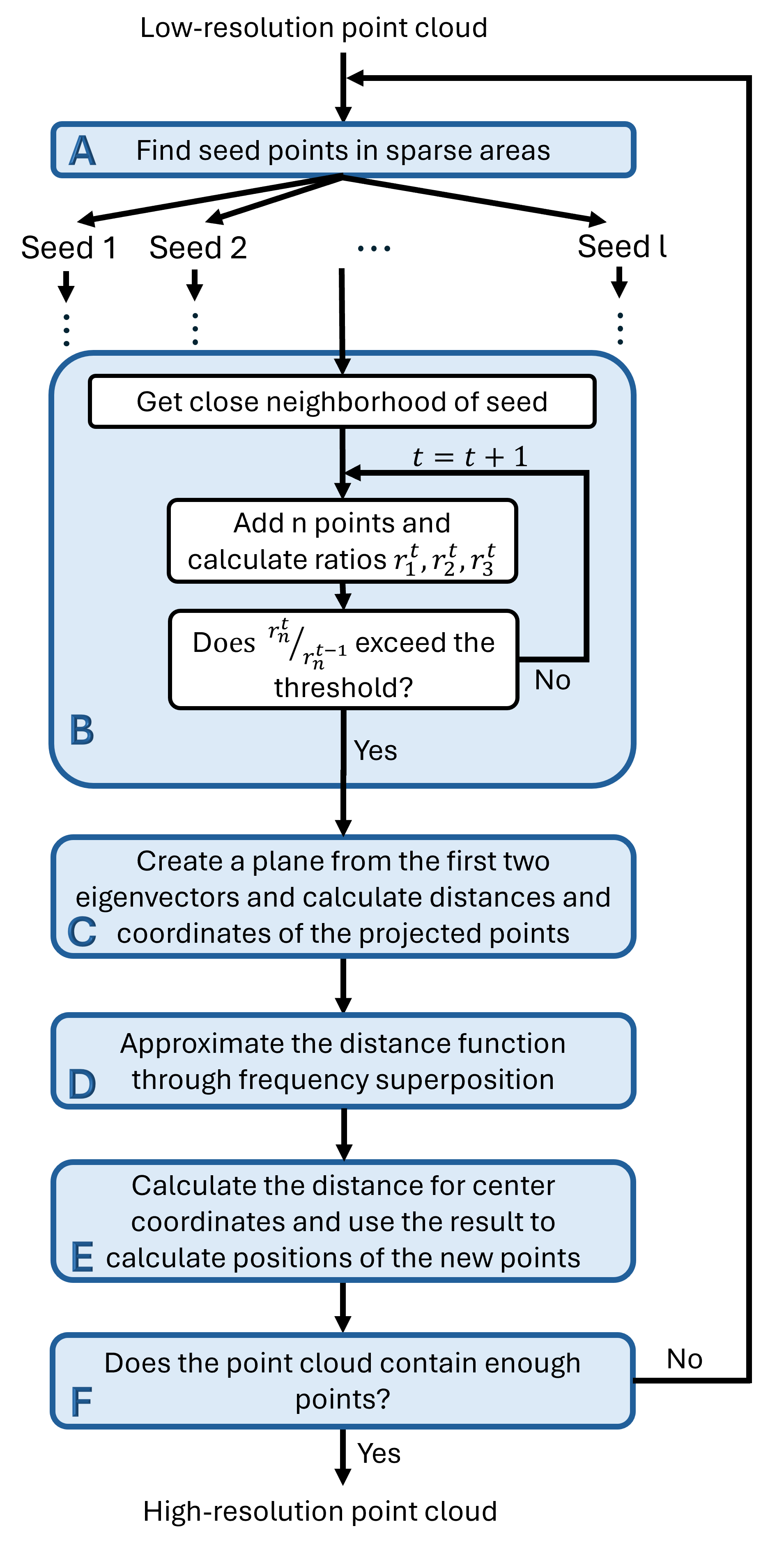}}
\caption{Flow chart of proposed method}
\label{fig:scheme}
\end{figure}

\noindent This work extends the principle of FSGU and explores the upsampling of point clouds through the superposition of multiple frequencies. In order to achieve reliable results with this approach, the partitioning of the point cloud and the model creation are crucial. Therefore, PUtPFS first identifies suitable clusters of points. \\
\indent The method can be divided into two parts. The first part is selecting a subset of the point cloud that will be considered as a patch to be upsampled. In the flow chart in Fig. \ref{fig:scheme}, this selection corresponds to blocks A and B. The second part is concerned with transforming the points in a way that we can apply the frequency selective method (block C), do the approximation of the surface (block D) and add new points to the patch (block E). The method can stop at any time when the desired number of points is reached (block F). Therefore, as an advantage, any upsampling factor can be chosen.\\

\begin{figure*}[t]
\centering
\centerline{\includegraphics[height=5.2cm]{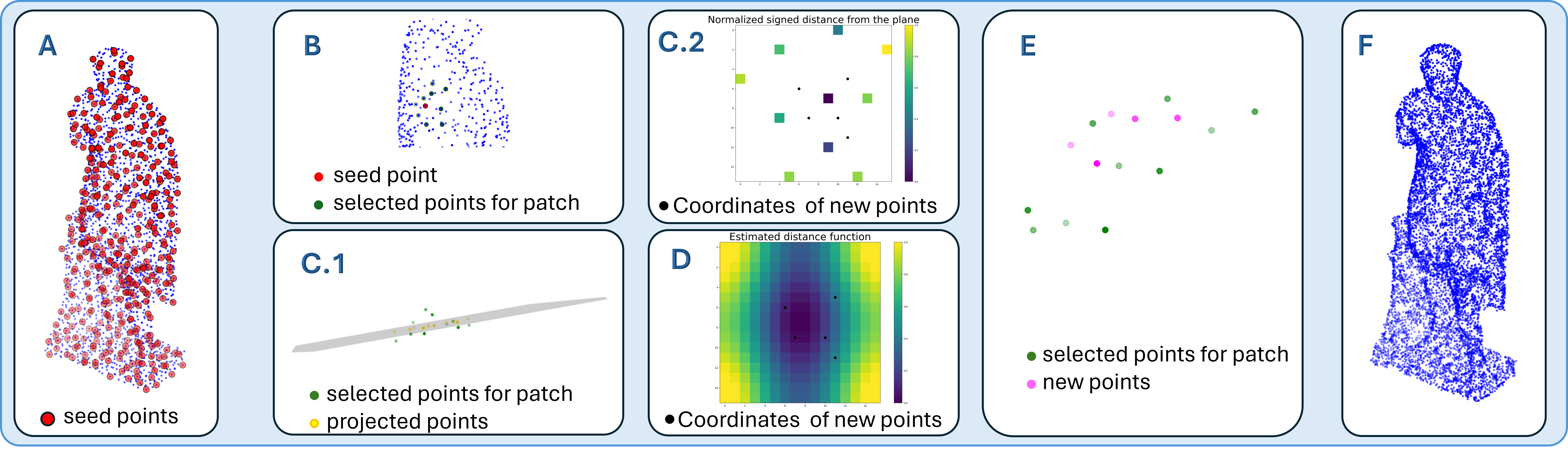}}
\caption{Visualization of the main steps of PUtPFS}
\label{fig:patch}
\end{figure*}

\subsection{Seed Points Selection}
\label{ssec:subhead}
\noindent The final result should be a uniformly upsampled point cloud. To this end, we want to get a good selection of points that are then used as seed points for the regions to be upsampled in the current round. We therefore start by sorting the points by the density of their neighborhood. This is done by comparing the average distance of the points to their $z$ closest neighbors. Then, starting with the point with the highest distance average, the points are successively added to the seed point set. However, the point is only chosen as a seed point if none of his $z$ nearest neighbors are already seed points. With this technique, we achieve a uniformly distributed set of seed points that focuses on sparse regions. An example of seed points in the first upsampling round of our approach can be seen on the left side in Fig. \ref{fig:patch}. For each region, the patch is then initially formed by the seed point and its $j_{1}$ nearest neighbors.\\

\subsection{Growing Patches}
\label{ssec:patches}
\noindent For the patch selection, we perform eigenvalue decompositions on subsets of points in the point cloud \cite{pcinterpret}. Looking at a point $\boldsymbol{\mathit{x_{0}}}$ and its $k$ closest neighbors $\boldsymbol{\mathit{x}_{i}}, i=1,...,k$, we can calculate a 3D covariance matrix $\boldsymbol{\mathit{S}}$
\begin{equation}
    \boldsymbol{\mathit{S}}= \frac{1}{k+1}\sum_{i=0}^{k}(\boldsymbol{\mathit{x}_{i}}-\boldsymbol{\mathit{\bar{x}}})(\boldsymbol{\mathit{x}_{i}}-\boldsymbol{\mathit{\bar{x}}})^{T}.
\end{equation}
Here, the variable $\boldsymbol{\mathit{\bar{x}}}$ is the geometric center, which is defined as
\begin{equation}
    \boldsymbol{\mathit{\bar{x}}} = \frac{1}{k+1}\sum_{i=0}^{k}\boldsymbol{\mathit{x}_{i}}.
\end{equation}
\noindent Because $\boldsymbol{\mathit{S}}$ is a symmetric positive-definite matrix, we have three eigenvalues $\lambda_{1}, \lambda_{2}, \lambda_{3}$ that correspond to an orthogonal system of eigenvectors. The eigenvalues can be used to describe the structure of the points. If, for example, $\lambda_{1} >> \lambda_{2}, \lambda_{3}$, the points show a mostly linear structure, whereas $\lambda_{1} \approx \lambda_{2} >> \lambda_{3}$ suggests a planar structure. If we grow a patch by continuously adding more nearest neighbors of a center point, we can define a stopping criterion for the growth by looking at the change of the ratios of the eigenvalues. In case adding a few points leads to a significant change, the points that were added might not belong to the same surface patch in the point cloud, or the surface is shaped in a way that makes it hard to describe it with a single model. Then, to obtain good surface estimations, patch growth is terminated before the last added points.\\
In our method, the patch region is iteratively expanded by adding $j_{2}$ points at each step until our predefined stopping criterion is met. We calculate the eigenvalues and the three ratios
\begin{equation}
    r_{1} = \frac{\lambda_{1}}{\lambda_{2}},\;\;
    r_{2} = \frac{\lambda_{1}}{\lambda_{3}},\;\;
    r_{3} = \frac{\lambda_{2}}{\lambda_{3}}
\end{equation}
of the current point patch at every step. In the following, the change of these ratios that occurred as a result of the additional points in the patch is computed with
\begin{equation}
    \Delta r_{n} = \frac{r_{n}^{t}}{r_{n}^{t-1}}.
\end{equation}
As stopping criterion, we use 
\begin{equation}
    |\Delta r_{n}-1| \geq P
\end{equation}
as a thresholding function. If the condition is true for any one of the three ratios, the patch growth stops and the previous point selection is chosen as a patch. An example of a selected patch can be seen in Fig. \ref{fig:patch} in block B.\\

\subsection{Point Transformations}
\label{sssec:transforming}
\noindent After the selection of the patch, a local upsampling shall be performed on the sparse points. Therefore, we approximate the surface by a parametrized function and add new points on this function. To this end, we need to describe a three-dimensional surface by a two-dimensional function, where the position in one dimension is given as the function value of the other two. For that reason, a plane is constructed from the eigenvectors of the two larger eigenvalues of the patch, and the selected points are projected onto this plane. This is visualized in block C.1 in Fig. \ref{fig:patch}. Now, we can interpret the signed distance of the points from the plane as a function of the position of their projected points. Therefore, we subdivide the area of the plane that encompasses the projected points into a grid.\\
\indent In the next step, we determine the grid coordinates for the new points. For this, we perform a Delaunay triangulation and use the centroids of the triangles as candidates for the coordinates of new points. The surface approximation tends to be less reliable near the patch boundaries. For this reason, only coordinate candidates that are not in the first or last $y\:\%$ of rows or columns in our patch are selected. The final grid coordinates for the new points can be seen in Fig. \ref{fig:patch} in blocks C.2 and D as black dots. The former shows the grid with the distance values from the original points. The latter visualizes the approximation of the function of the surface.\\

\begin{table*}[t]
\caption{Quantitative comparison on the PU-GAN and PU1K datasets for an upsampling factor of four.}
\label{Tab: result}
\begin{center}
\begin{tabular}{|c||c|c|c||c|c|c||c|c|c||c|c|c|}
\hline
& \multicolumn{6}{|c||}{PU-GAN}&\multicolumn{6}{|c|}{PU1K}\\
\hline
& \multicolumn{3}{|c||}{input of 1024 points}&\multicolumn{3}{|c||}{input of 2048 points}& \multicolumn{3}{|c||}{input of 1024 points}&\multicolumn{3}{|c|}{input of 2048 points}\\
\textbf{Method} & CD & HD & P2F & CD & HD & P2F & CD & HD & P2F & CD & HD & P2F\\
 & \scriptsize$\times10^{-3}$ & \scriptsize$\times10^{-3}$ & \scriptsize$\times10^{-3}$ & \scriptsize$\times10^{-3}$ & \scriptsize$\times10^{-3}$ & \scriptsize$\times10^{-3}$& \scriptsize$\times10^{-3}$ & \scriptsize$\times10^{-3}$ & \scriptsize$\times10^{-3}$ & \scriptsize$\times10^{-3}$ & \scriptsize$\times10^{-3}$ & \scriptsize$\times10^{-3}$\\
\hline

\footnotesize PU-Net \cite{PUNET} & \footnotesize 1.309 & \footnotesize 19.138 & \footnotesize 11.970 & \footnotesize 0.817 & \footnotesize 11.150 & \footnotesize 7.838 & \footnotesize 1.899 & \footnotesize 24.754 & \footnotesize 7.321 & \footnotesize 1.155 & \footnotesize 15.170 & \footnotesize 4.847\\
\footnotesize MPU \cite{MPU}  & \footnotesize 1.236 & \footnotesize 16.116 & \footnotesize 8.449 & \footnotesize 0.713 & \footnotesize 10.614 & \footnotesize 5.381 & \footnotesize 1.679 & \footnotesize 21.119 & \footnotesize 5.450 & \footnotesize 0.935 & \footnotesize 13.327 & \footnotesize 2.963 \\
\footnotesize PU-GAN \cite{PUGAN} & \footnotesize 0.768 & \footnotesize 12.250 & \footnotesize 6.593 & \footnotesize 0.469 & \footnotesize 8.220 & \footnotesize 4.047 & \footnotesize 1.132 & \footnotesize 14.809 & \footnotesize 4.530 & \footnotesize 0.707 & \footnotesize 10.411 & \footnotesize 2.963 \\
\footnotesize PU-GCN \cite{PUGCN} & \footnotesize 0.774 & \footnotesize 9.594 & \footnotesize 6.197 & \footnotesize 0.401 & \footnotesize 5.630 & \footnotesize 3.650 & \footnotesize 1.035 & \footnotesize 12.032 & \footnotesize 3.946 & \footnotesize 0.585 & \footnotesize 7.577 & \footnotesize 2.504 \\
\footnotesize SPU-Net \cite{SPUNET} & \footnotesize 1.104 & \footnotesize 39.023 & \footnotesize 10.289 & \footnotesize 0.509 & \footnotesize 23.497 & \footnotesize 6.106 & \footnotesize 1.338 & \footnotesize 37.368 & \footnotesize 6.444 & \footnotesize 0.955 & \footnotesize 21.054 & \footnotesize 4.083 \\
\footnotesize SPU-PMD \cite{SPUPMD} & \footnotesize \textbf{0.602} & \footnotesize \textbf{5.762} & \footnotesize 4.040 & \footnotesize \textbf{0.314} & \footnotesize 3.320 & \footnotesize 2.441 & \footnotesize \textbf{0.892} & \footnotesize 8.252 & \footnotesize 2.765 & \footnotesize 0.544 & \footnotesize 4.926 & \footnotesize 1.861 \\
\hline
\footnotesize EAR \cite{EAR} & \footnotesize 1.046 & \footnotesize 7.389 & \footnotesize 8.117 & \footnotesize 0.477 & \footnotesize 3.580 & \footnotesize 4.615 & \footnotesize 1.301 & \footnotesize 10.385 & \footnotesize 5.329 &  \footnotesize 0.751 & \footnotesize 5.924 & \footnotesize 3.607 \\
\footnotesize FSGU \cite{FSGU} & \footnotesize 1.729 & \footnotesize 39.114 & \footnotesize 537.985 & \footnotesize 1.744 & \footnotesize 15.42 & \footnotesize 541.132 & \footnotesize 3.517 & \footnotesize 77.174 & \footnotesize 302.239 & \footnotesize 1.720 & \footnotesize 33.774 & \footnotesize 303.512 \\
\hline
\footnotesize Ours & \footnotesize \underline{0.750} & \footnotesize \underline{6.944} & \footnotesize \textbf{\underline{3.461}} & \footnotesize \underline{0.368} & \footnotesize \underline{\textbf{3.168}} & \footnotesize \textbf{\underline{2.025}} & \footnotesize \underline{0.924} & \footnotesize \textbf{\underline{8.068}} & \footnotesize \textbf{\underline{2.368}} & \footnotesize \textbf{\underline{0.533}} & \footnotesize \textbf{\underline{4.693}} & \footnotesize \textbf{\underline{1.536}}  \\
\hline
\end{tabular}
\label{tab1}
\end{center}
\end{table*}

\subsection{Upsampling of Selected Area}
\label{sssec:upsamling}
\noindent For the approximation, we use the frequency selective method, which is a well-performing algorithm for image reconstruction and resampling \cite{FSR} \cite{FSMR}. This algorithm exploits the fact that image signals have a sparse representation in the frequency domain. With the known samples of a function, the method generates $g[m,n]$, an approximation of the signal, as a weighted superposition of the basis functions. In
\begin{equation}
    g[m,n] = \sum_{(k,l)\in K} \hat{c}_{(k,l)}\varphi_{(k,l)}[m,n],
\end{equation}
the variable $\varphi_{(k,l)}[m,n]$ represents the basis functions. In our method, we use the discrete cosine transform as orthogonal bases. The set $K$ contains all the basis functions used to approximate the signal in the area $\boldsymbol{\mathit{A}} = \{(m, n)\in \mathbb{Z}^{2}|0\leq m < M, 0 \leq n <N \}$. The expansion coefficients $\hat{c}_{(k,l)}$ must be determined. Therefore, an iterative procedure is proposed. The initial model $g^{(0)}[m,n]$ is set to zero, and in each iteration $v$ an additional basis function $\varphi_{(u,w)}[m,n]$ is added as
\begin{equation}
    g^{(v)}[m,n]=g^{(v-1)}[m,n]+\hat{c}_{(u,w)}\varphi_{(u,w)}^{v}[m,n].
\end{equation}
Similarly to \cite{FSMR}, the coefficient $\hat{c}^{(v)}$ is chosen such that the weighted residual energy $E^{v}$ is minimized. In each iteration, the basis function is selected to maximize the difference between the current and the previous residual energy value. We therefore want to choose $(u,w)$ as
\begin{equation}
    (u,w) = \underset{(k,l)}{\mathrm{argmax}}\, \Delta E_{(k,l)}^{(v)}.
\end{equation}
The number of iterations depends on the number of points in the patch with a higher amount of points leading to more iterations. We can achieve better results by additionally adding a spatial and a frequency weighting. Both weightings use a radially symmetric exponential weighting function defined as
\begin{equation}
    w_{s}(d_{s}) = \phi_{1}^{d_{s}} \quad \text{and} \quad w_{f}(d_{f}) = \phi_{2}^{d_{f}}
\end{equation}
The parameters $\phi_{1}$ and $\phi_{2}$ determine the decay rates. The spatial weights $w_{s}$ are calculated with the Euclidean distance from the grid center $d_{s}$.  The weighting of our two-dimensional DCT basis functions makes the method focus on lower frequencies by assigning a weight $w_{f}$, which decays with its distance from the origin in the frequency domain $d_{f}$.\\
\subsection{Adding New Points}
\label{sssec:newpoints}
\noindent After approximating the function, we use our result to find the positions of new points. For this, we check the function values of the grid coordinates of our new points and invert the transform that was performed in block C for those coordinates. The calculated positions for the additional points can be seen in magenta in block E of Fig. \ref{fig:patch}.\\
\indent With the steps described in C, D, and E, the patches of each seed point are densified. Subsequently, our algorithm compares the current number of points in the point cloud with the desired amount. If the point cloud does not contain enough points, new seed points are chosen, otherwise, the algorithm terminates.\\

\begin{figure}[t]
\centering
\centerline{\includegraphics[height=5.5cm]{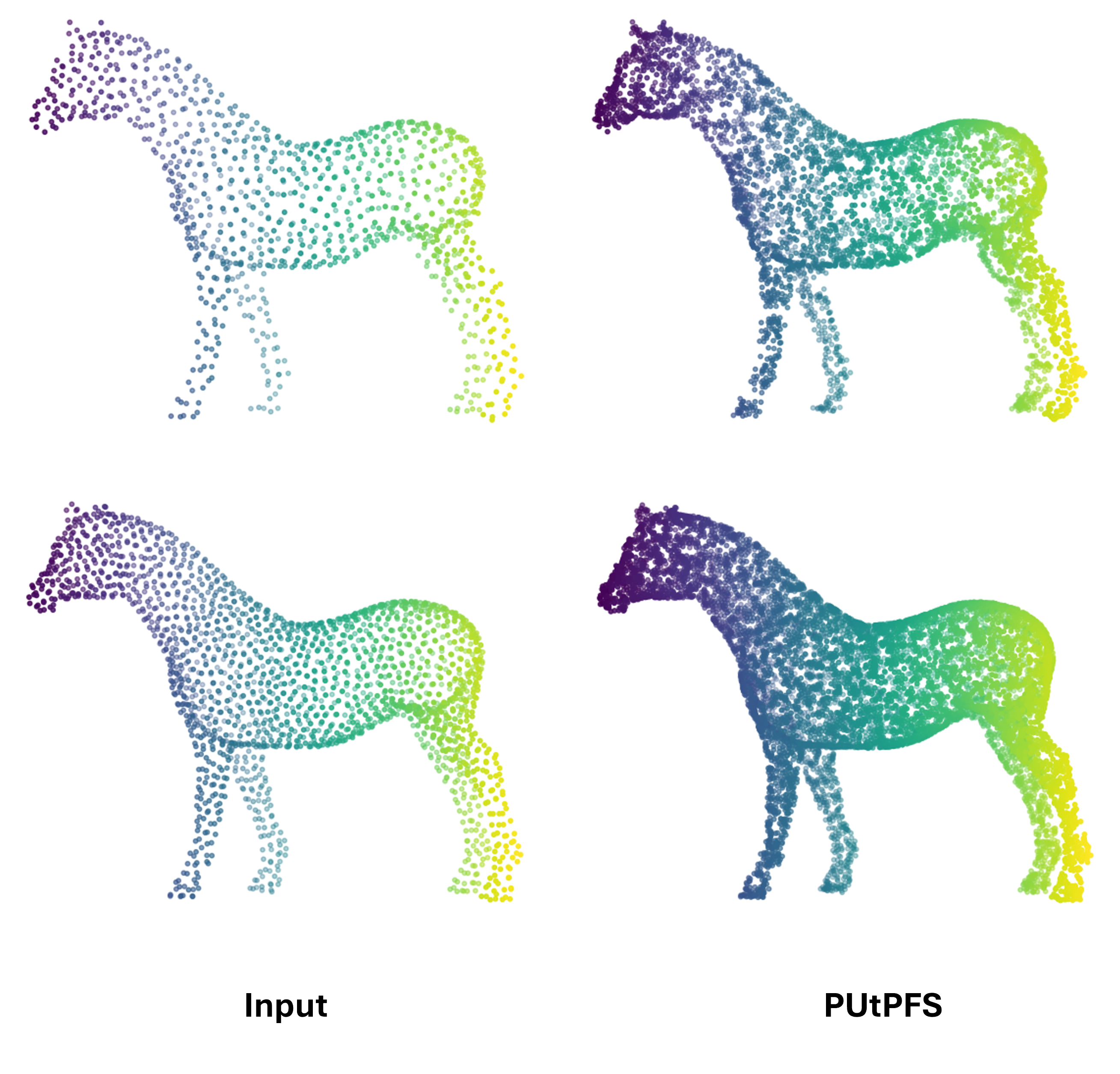}}
\caption{Visual example of two input sizes and the upsampling results}
\label{fig:horse}
\end{figure}

\section{PERFORMANCE EVALUATION}
\label{sec:eval}
\begin{figure}[t]
\centering
\centerline{\includegraphics[height=5.5cm]{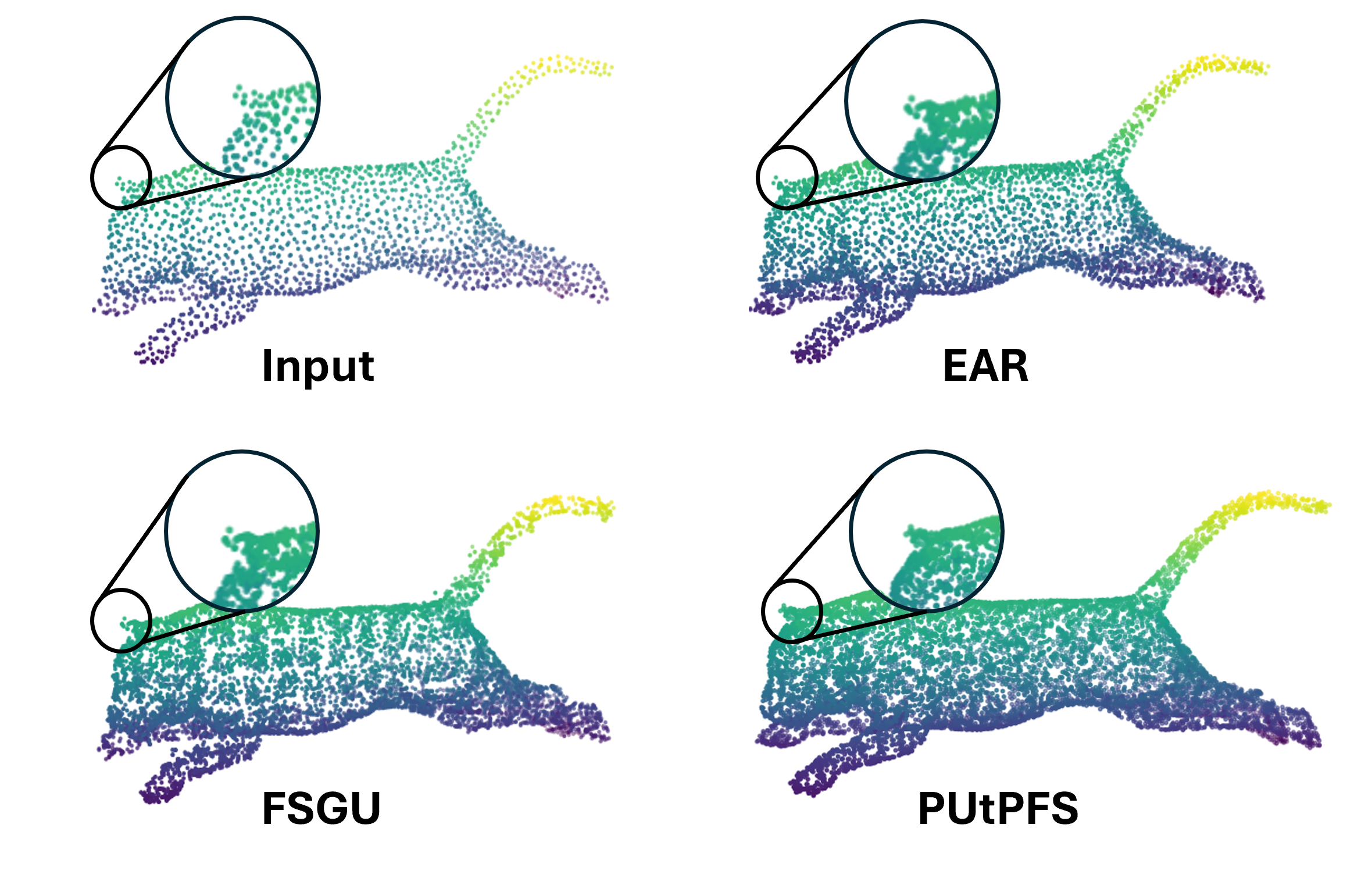}}
\caption{Comparison of upsampling results of the optimization-based approaches}
\label{fig:tiger}
\end{figure}

\noindent \textbf{Datasets.} This section compares PUtPFS with state-of-the-art upsampling methods. As in most other point cloud upsampling publications, we evaluate the effectiveness on the two synthetic datasets PU1K and PU-GAN. Both datasets contain training and test data. For a fair comparison with the deep learning-based approaches, we only average the result over the test sets, which include 27 point clouds in the PU-GAN and 127 point clouds in the PU1K dataset. In Fig. \ref{fig:horse} we present one visual example of our upsampling with the upsampling factor of four. The top row shows an input point cloud containing 1024 points and its upsampled result. The bottom row has 2048 points as input. The advantage of using the two synthetic datasets lies in the availability of ground-truth geometry, which enables an accurate calculation of performance metrics.\\
\indent \textbf{Parameter Settings.} For our seed points, we check the average distance to the $z=10$ closest neighbors for all points. For our patches, we set $j_{1}$ to 4 and $j_{2}$ to 2. For stopping the patch growth, we use $P=0.5$ in the first two iterations and $P=0.1$ subsequently. To avoid huge patch sizes that lead to a slow frequency selective method, the maximum range of patch growth iterations is set to $20$. After the plane is fitted to the patch, the area of the plane that encompasses the projections of the patch points is divided into a grid with a resolution of 16x16. To choose the plane coordinates for the new points, we set $y = 25 \:\%$. In the approximation of the function, we use the weighting parameters $\phi_1 = \phi_{2} = 0.88$.\\
\indent \textbf{Evaluation.} A quantitative evaluation can be seen in Table \ref{Tab: result} for both datasets. Following other recent works, we calculate the Chamfer Distance (CD), Hausdorff distance (HD) and point-to-surface distance (P2F). For all metrics, a smaller number indicates a better result. We compare our method with the state-of-the-art networks PU-Net \cite{PUNET}, MPU \cite{MPU}, PU-GAN \cite{PUGAN}, PU-GCN \cite{PUGCN}, SPU-Net \cite{SPUNET}, SPU-PMD \cite{SPUPMD} and the two optimization-based approaches EAR \cite{EAR} and FSGU \cite{FSGU}. The best results are marked in bold, and underlined numbers indicate the best value among the optimization-based approaches. All methods perform worse in the more challenging PU1K dataset compared to PU-GAN. For both sets, our method achieves a CD and HD, which are in the same range as the deep learning-based approaches. In some cases we perform better and in some cases slightly worse than the best performing architecture. However, when considering the P2F, PUtPF consistently gives the best results. Looking only at the other two optimization-based approaches, PUtPFS is outperforming both of them. In Fig. \ref{fig:tiger} a visual comparison is given between the upsampling results of the three optimization-based approaches. PUtPFS is less noisy and the interpolated points better follow the surface. This can be seen, for example, in the ear region of the tiger.\\
\indent \textbf{Limitations.} While the explainability and complete independence of trainings data are big advantages of our method, it also inherits some drawbacks. For one, PUtPFS is not able to correctly upsample all possible types of point cloud structures. Two very close surfaces might not be interpolated in a sensible way and there is a limitation on thin line structures which cannot be well approximated by 2D surfaces. Secondly, the point placement is not perfectly uniform.\\
\indent \textbf{Complexity.} Compared with the next best optimization-based approach, EAR, we need $47\:\%$ of their inference time. The complexity of PUtPFS strongly depends on the structure of the point cloud, which influences the patch size and the iterations of the surface approximation. However, if we assume a similar structure and size of patches, our inference time scales linearly with the number of input points.\\

\section{CONCLUSION}
\label{sec:conclusion}
\noindent This paper presents a novel point cloud upsampling method which interpolates missing samples using an initial patch growing and subsequent frequency selective surface approximation. With this strategy, PUtPFS manages to achieve state-of-the-art results and exceeds the results of current deep learning-based methods in the P2F metric on the common test datasets PU-GAN and PU1K. Additionally, PUtPFS does not need training data and is not a black-box approach, which is important in some areas, for example the medical field. If we compare our method with other optimization-based approaches, we also achieve the best results in the CD and HD metrics.


\begin{thebibliography}{00}
\bibitem{reviewPaper} Yan Zhang, Wenhan Zhao, Bo Sun, Ying Zhang, and Wen Wen, “Point cloud upsampling algorithm: A systematic review,” Algorithms, vol. 15, no. 4, 2022.
\bibitem{PUNET} Lequan Yu, Xianzhi Li, Chi-Wing Fu, Daniel Cohen-Or, and Pheng-Ann Heng, “ PU-Net: Point Cloud Up-sampling Network ,” in 2018 IEEE/CVF Conference on Computer Vision and Pattern Recognition (CVPR), Los Alamitos, CA, USA, June 2018, pp. 2790–2799, IEEE Computer Society.
\bibitem{Pointnet++} Charles R. Qi, Li Yi, Hao Su, and Leonidas J. Guibas, “Pointnet++: deep hierarchical feature learning on point sets in a metric space,” in Proceedings of the 31st International Conference on Neural Information Processing Systems, Red Hook, NY, USA, 2017, NIPS’17, p.5105–5114, Curran Associates Inc.
\bibitem{PUGAN} Ruihui Li, Xianzhi Li, Chi-Wing Fu, Daniel Cohen-Or, and Pheng-Ann Heng, “PU-GAN: A Point Cloud Upsampling Adversarial Network,” in 2019 IEEE/CVF International Conference on Computer Vision (ICCV), Los Alamitos, CA, USA, Nov. 2019, pp. 7202–7211, IEEE Computer Society.
\bibitem{PUGCN} Guocheng Qian, Abdulellah Abualshour, Guohao Li, Ali Thabet, and Bernard Ghanem, “PU-GCN: Point Cloud Upsampling using Graph Convolutional Networks,” in 2021 IEEE/CVF Conference on Computer Vision and Pattern Recognition (CVPR), Los Alamitos, CA, USA, June 2021, pp. 11678–11687, IEEE Computer Society.
\bibitem{GCPCU} Dandan Ding, Chi Qiu, Fuchang Liu, and Zhigeng Pan, “Point cloud upsampling via perturbation learning,” IEEE Transactions on Circuits and Systems for Video Technology, vol. 31, no. 12, pp. 4661–4672, 2021.
\bibitem{PointTrans}Nico Engel, Vasileios Belagiannis, and Klaus Dietmayer, “Point transformer,” IEEE Access, vol. 9, pp.134826–134840, 2021.
\bibitem{SPUNET} Xinhai Liu, Xinchen Liu, Yu-Shen Liu, and Zhizhong Han, “Spu-net: Self-supervised point cloud upsampling by coarse-to-fine reconstruction with self-projection optimization,” Trans. Img. Proc., vol. 31, pp. 4213–4226, Jan. 2022.
\bibitem{SPUPMD} Yanzhe Liu, Rong Chen, Yushi Li, Yixi Li, and Xuehou Tan, “Spu-pmd: Self-supervised point cloud upsampling via progressive mesh deformation,” in 2024 IEEE/CVF Conference on Computer Vision and Pattern Recognition (CVPR), 2024, pp. 5188–5197.
\bibitem{LOP} Yaron Lipman, Daniel Cohen-Or, David Levin, and Hillel Tal-Ezer, "Parameterization-free projection for geometry reconstruction,” ACM Trans. Graph., vol. 26, no. 3, pp. 22–es, July 2007.
\bibitem{WLOP} Reinhold Preiner, Oliver Mattausch, Murat Arikan, Renato Pajarola, and Michael Wimmer, “Continuous projection for fast l1 reconstruction,” ACM Trans. Graph., vol. 33, no. 4, July 2014.
\bibitem{CLOP} Hui Huang, Dan Li, Hao Zhang, Uri Ascher, and Daniel Cohen-Or, “Consolidation of unorganized point clouds for surface reconstruction,” ACM Trans. Graph., vol. 28, no. 5, pp. 1–7, Dec. 2009.
\bibitem{EAR} Hui Huang, Shihao Wu, Minglun Gong, Daniel Cohen-Or, Uri Ascher, and Hao (Richard) Zhang, “Edge-aware point set resampling,” ACM Trans. Graph., vol. 32, no. 1, Feb. 2013.
\bibitem{FSGU} Viktoria Heimann, Andreas Spruck, and André Kaup, “Frequency-Selective Geometry Upsampling of Point Clouds,” in IEEE International Conference on Image Processing (ICIP) 2022, 2022.
\bibitem{pcinterpret} Martin Weinmann, Boris Jutzi, Stefan Hinz, and Clément Mallet, “Semantic point cloud interpretation based on optimal neighborhoods, relevant features and efficient classifiers,” ISPRS Journal of Photogrammetry and Remote Sensing, vol. 105, 02 2015.
\bibitem{FSR} Jürgen Seiler, Markus Jonscher, Michael Schoberl, and André Kaup, “Resampling images to a regular grid from a non-regular subset of pixel positions using frequency selective reconstruction,” Trans. Img. Proc., vol. 24, no. 11, pp. 4540–4555, Nov. 2015.
\bibitem{FSMR} Ján Koloda, Jürgen Seiler, and André Kaup, “Frequency-selective mesh-to-grid resampling for image communication,” IEEE Transactions on Multimedia, vol. 19, no. 8, pp. 1689–1701, 2017.
\bibitem{MPU} Wang Yifan, Shihao Wu, Hui Huang, Daniel Cohen-Or, and Olga Sorkine-Hornung, "Patch-based progressive 3d point set upsampling,” in Proceedings of the IEEE/CVF Conference on Computer Vision and Pattern Recognition (CVPR), June 2019.

\end{thebibliography}
\end{document}